\documentclass[journal]{IEEEtran}
\usepackage{cite}

\usepackage[T1]{fontenc}
\usepackage{paralist}
\usepackage{mathptmx}   
\usepackage{caption}
\usepackage{amsmath,amssymb,amsfonts}
\usepackage{amsthm}
\usepackage{mathtools}
\usepackage{array}
\usepackage{bm}
\usepackage{booktabs}
\usepackage{multirow}
\usepackage{multicol}
\usepackage{cases}
\usepackage{algorithm}
\usepackage{algorithmic}
\usepackage{graphicx}
\usepackage{subfigure}               
\usepackage[colorlinks=true, 
            linkcolor=blue, 
            citecolor=blue,
            urlcolor=blue]{hyperref}

\begin{document}
%
%
%
\title{Great X: A Unified Multi-Modal Simulator Bridging the Sim2Real Gap for 6G}
\author{Yuan~Gao,~\IEEEmembership{Member,~IEEE}, Kongwu~Huang, Jun~Jiang, Shiyi~Mu, Shugong~Xu,~\IEEEmembership{Fellow,~IEEE}

\thanks{This work was supported in part by Shanghai Natural Science Foundation under Grant 22ZR1422200, and in part supported by the 6G Science and Technology Innovation and Future Industry Cultivation Special Project of Shanghai Municipal Science and Technology Commission under Grant 24DP1501001. Corresponding authors: Shugong Xu.}
\thanks{Yuan Gao, Kongwu Huang and Shiyi Mu are with Shanghai University, Shanghai, China, e-mail: gaoyuansie@shu.edu.cn, huangkongwu@shu.edu.cn and shiyimu@shu.edu.cn.}%
\thanks{Jun Jiang and Shugong Xu are with the School of Advanced Technology, Xi'an Jiaotong-Liverpool University, Suzhou, China, e-mail: Jun.Jiang25@student.xjtlu.edu.cn and shugong.xu@xjtlu.edu.cn.}}%
\maketitle
\begin{abstract}
Large-scale, precisely time-synchronized multi-modal datasets are essential for data-driven research toward sixth-generation (6G) wireless networks in complex three-dimensional environments, yet real-world measurements remain prohibitively difficult. Existing simulators rely on multi-platform co-simulation frameworks that introduce timing misalignment between processes, support only narrow sets of modalities, and employ oversimplified geometric primitives with static material parameters, all of which degrade fidelity and widen the simulation-to-reality (Sim2Real) gap. To this end, we propose Great-X, which resolves these issues through a fully unified single-engine design implemented inside Unreal Engine. The electromagnetic propagation pipeline was rebuilt around the engine's native ray-tracing functions, while physically based rendering materials are directly co-located with electromagnetic properties on the same 3D meshes. This produces pixel-level spatial consistency between visual and radio outputs. A deterministic fixed time-step clock together with synchronous buffering across heterogeneous sensors eliminates latency jitter and clock drift, delivering frame-accurate alignment among channel state information (CSI), red-green-blue (RGB) images, Light Detection and Ranging (LiDAR) point clouds, radar, and event streams. Using this architecture we created the Great multi-modal communication dataset (Great-MCD), which contains more than three million temporally and spatially aligned samples collected across urban and rural scenes, day and night lighting, multiple unmanned aerial vehicle types, and both straight and curved trajectories. By addressing the fundamental synchronization, fidelity, and scalability shortcomings of conventional multi-platform approaches, Great-X supplies high-fidelity synthetic data that helps bridge the Sim2Real gap in next-generation 6G wireless systems. Extensive simulations validate the fidelity of the data generated by Great-X. The channel impulse responses closely match those obtained from real measurements, and the data enable models that achieve better zero-shot and fine-tuned performance in wireless positioning and CSI feedback tasks than those trained on simulated data from existing simulators or directly on measured data.
\end{abstract}


\IEEEpeerreviewmaketitle
\section{Introduction}

\IEEEPARstart{T}he evolution of sixth-generation (6G) wireless networks toward AI-native architectures has shifted the research paradigm from traditional model-based analysis and optimization toward data-driven artificial intelligence (AI) methodologies~\cite{chen20235g,gao2025enabling,wang2023road}. This is becoming the dominate research direction in channel estimation~\cite{OTFSGao2025}, channel extrapolation~\cite{SSnet2025gao,jin2025linformer}, resource allocation, wireless positioning~\cite{xu2025enhanced}, beam management~\cite{jin2026generalizable}, integrated sensing and communication~\cite{aldirmaz2025comprehensive}, etc. While model mismatch between idealized analytical assumptions and real-world imperfections has long existed in wireless communications~\cite{gao2026csiextra,elzanaty2024near,newport2007experimental}, the rise of deep learning has significantly amplified the simulation-to-reality (Sim2Real) gap due to scarcity of the practical dataset~\cite{ruah2025bridge}. Data-driven models, with their strong capacity to fit complex patterns, tend to overfit simulation-specific artifacts—such as simplified material properties, perfect synchronization, and static scattering statistics—leading to severe performance degradation when deployed on real measurements~\cite{muruganandham2025smart}. This challenge is especially pronounced in dynamic three-dimensional environments with heterogeneous sensors, for example in low-altitude unmanned aerial vehicle (UAV) operations~\cite{alkhateeb2023deepsense}. In such scenarios, acquiring large-scale, temporally and spatially aligned multi-modal datasets that jointly capture electromagnetic propagation and environmental perception remains extremely difficult due to hardware costs, regulatory constraints, and inherent cross-sensor clock drift~\cite{charan2025sensing}. Consequently, models trained on limited or misaligned real-world data suffer from pronounced domain shift, undermining generalization and practical deployment~\cite{feng2023dda}.

Existing simulation platforms have sought to alleviate data scarcity by integrating wireless channel simulators with graphics engines or autonomous-driving platforms~\cite{borges2024caviar,alkhateeb2023deepsense}. However, these multi-engine, loosely coupled architectures inherently suffer from timing misalignment, inter-process communication latency, and inconsistent spatial modeling between radio propagation and visual semantics~\cite{hauweele2019toward}. Scene representations are typically oversimplified, relying on coarse geometric primitives and static material parameters that fail to capture the fine-grained, frequency-dependent scattering and occlusion effects present in real environments~\cite{ruah2025bridge}. Although recent hybrid model-driven and data-driven approaches attempt to mitigate some of these issues by embedding physical priors, they still depend heavily on the fidelity and synchronization quality of the underlying simulation data~\cite{liang2024data,jin2023model}. As a result, synthetic datasets generated by current tools exhibit limited transferability to real measurements, leaving the amplified Sim2Real gap in data-driven 6G systems largely unaddressed~\cite{muruganandham2025smart}.

\begin{table*}[tbp]
\centering
\small
\caption{Comparison of representative multi-modal ISAC frameworks.}
\begin{tabular}{cccc}
\toprule
{Name} & {\# Engines} & {Architecture} &  {Modalities Supported} \\ 
\midrule
DeepVerse-6G \cite{DeepVerse} & 2 & Hybrid  & CSI, Depth, LiDAR, RGB, Radar \\
SynthSoM \cite{Cheng2025SynthSoM} & 3 & Hybrid  & CSI, Depth, LiDAR, RGB, Radar \\
\textbf{Great-X (Ours)} & \textbf{1} & \textbf{Unified} &  CSI, Depth, LiDAR, RGB, Radar, \textbf{Event} \\
\bottomrule
\end{tabular}
\label{tab:sim_compare}
\end{table*}

Future 6G systems are expected to move well beyond the traditional focus on higher data rates, lower latency, and massive connectivity. Instead, they are envisioned to become intelligent platforms that can actively sense, interpret, and adapt to their surrounding physical environment in real time~\cite{tang2025cooperative,zeng2021toward}. This shift is driven by emerging applications such as low-altitude economy, UAV swarms, digital twins, autonomous transportation, and immersive extended reality, which demand not only reliable communication links but also a deep understanding of the dynamic physical world~\cite{song2025overview,tang2025cooperative}. In these scenarios, the network must perceive environmental changes (e.g., moving obstacles, varying user density, or material properties), predict their impact on radio propagation, and proactively adjust transmission strategies~\cite{wu2025isac}. Without such environment-aware capabilities, conventional connectivity-centric designs suffer from excessive overhead, poor adaptability in dynamic settings, and limited ability to optimize joint sensing and communication performance~\cite{li2023channel,liu2025channel}. Realizing this form of environment-intelligent communication will therefore require simulation platforms that can generate large quantities of multi-modal data, channel state information (CSI) together with red-green-blue (RGB) images, depth maps, Light Detection and Ranging (LiDAR) point clouds, etc., that are both pixel-level consistent and physically accurate inside richly detailed three-dimensional scenes~\cite{borges2024caviar,ruah2025bridge}. Only with such unified and high-fidelity data generation can learning algorithms develop robust joint representations of radio propagation and environmental structure that remain effective when transferred from simulation to real-world deployment~\cite{muruganandham2025smart}.

To address these challenges, this paper proposes Great-X, a high-fidelity unified multi-modal simulation platform specifically designed for 6G to bridge the Sim2Real Gap. The main contributions of this work are as follows:
\begin{enumerate}
\item 
Built on Unreal Engine \cite{unrealengine}, Great-X features a completely redesigned electromagnetic propagation pipeline that overcomes the core shortcomings of existing multi-platform tools. The platform incorporates native ray-tracing-based wireless channel modeling, which forms the foundation of its unified single-engine architecture. It supports the broadest range of modalities among current ISAC simulators, enabling efficient and fully synchronous generation of CSI, RGB, depth, LiDAR, radar, and event camera data. We further construct Great-MCD\footnote{Will be soon available in https://github.com/hkw-xg/Great-MCD} based on Great-X, the largest-scale multi-modal UAV sensing dataset for low-altitude scenarios to date. It features synchronized camera, wireless channel, and spatial coordinate data, with over 3 million samples that far exceed the scale of existing relevant datasets.

\item 
Unlike fragmented co-simulation approaches that suffer from asynchronous timing and clock drift, Great-X employs a deterministic fixed-step clock model together with a synchronous blocking mechanism across seven heterogeneous modalities. This architecture guarantees zero-latency jitter in multi-sensor timestamping. Moreover, by strictly co-locating physically based rendering (PBR) materials with electromagnetic properties on unified 3D meshes, Great-X achieves pixel-level spatial consistency between visual semantics and radio propagation. This design enables seamless joint modeling and evaluation of communication and multi-modal sensing tasks while simultaneously generating precisely synchronized data across CSI, RGB images, depth maps, LiDAR, radar, and event camera outputs, as shown in Fig.~\ref{fig:Great-X_test}.

\item Extensive experiments demonstrate Great-X's strong physical fidelity and practical value. The channel impulse responses it generates closely match real-world DICHASUS indoor measurements in both structure and amplitude patterns. In multi-modal three-dimensional UAV positioning tasks, the proposed multi-modal knowledge-distillation (MMKD) framework trained on Great-X data significantly outperforms baselines from Sionna, QD, and even models trained directly on measured data. CSI feedback experiments using CsiNet further confirm that Great-X produces highly realistic channel characteristics, delivering superior zero-shot and fine-tuned performance compared with other simulators.
\end{enumerate}

The remainder of this paper is organized as follows. Section~\ref{sec:related_work} reviews related work on 6G simulators and multi-modal ISAC datasets. Section~\ref{sec:greatx} introduces the proposed Great-X simulator, describing its unified single-engine architecture and workflow as well as the construction of the accompanying Great-MCD dataset. Section~\ref{sec:core_modules} details the core modules and key technical features of Great-X. Section~\ref{sec:experiments} presents the simulation settings together with comprehensive performance evaluation, including physical fidelity validation against real-world DICHASUS measurements, multi-modal three-dimensional UAV positioning, and CSI feedback tasks. Finally, Section~\ref{sec:conclusion} concludes the paper.

\section{Related Work}
\label{sec:related_work}
\subsection{ISAC Simulator}

Most mainstream wireless communication datasets are generated using either commercial ray-tracing tools, such as Wireless InSite \cite{wireless_insite}, or open-source simulators such as NVIDIA’s Sionna \cite{sionna}. Although these tools are widely adopted for communication modeling \cite{testolina2024boston,iye2025open}, they are limited to producing CSI and provide no support for multi-modal sensing data.

Recent studies have sought to overcome this limitation by integrating sensing capabilities into ISAC simulation platforms. A representative example is DeepVerse 6G \cite{DeepVerse}, which builds upon DeepMIMO \cite{deepmimo} through a co-simulation framework that combines multi-modal sensing data—such as images and LiDAR—with wireless channel simulation using Wireless InSite and the CARLA \cite{carla} autonomous driving simulator. While this approach improves simulation versatility, it does not support low-altitude UAV scenarios. In addition, its dual-engine design increases system complexity and makes real-time synchronization difficult.

Similarly, SynthSoM \cite{Cheng2025SynthSoM} integrates Wireless InSite and WaveFarer \cite{wavefarer_website} with the AirSim \cite{airsim} platform to generate multi-modal datasets that include images, radar, and point clouds. Despite its broader modality coverage, SynthSoM still depends on a heterogeneous multi-engine architecture and relies on the commercial Wireless InSite software, which limits openness and scalability. 

As shown in Table~\ref{tab:sim_compare}, compared with SynthSoM and DeepVerse 6G, our simulation pipeline achieves a fully end-to-end process within a single Unreal Engine environment, eliminating dependence on commercial tools. Moreover, the modeling accuracy of these earlier platforms in complex or dynamic environments falls short of the high-fidelity, graphics-driven capabilities offered by Unreal Engine. In contrast, Great-X introduces a unified, single-engine, multi-modal simulation platform built entirely on Unreal Engine. It supports the synchronized generation of diverse modalities and newly incorporates event camera data.

\subsection{Multi-modal ISAC Dataset}
Existing ISAC datasets also suffer from notable limitations. DeepMIMO \cite{deepmimo} relies exclusively on Wireless InSite for channel simulation. While suitable for standardized communication modeling, it offers no support for user-defined environments and provides only single-modality CSI data, with no native integration of multi-modal sensing. The ViWi dataset \cite{viwi} is a synthetic multi-modal dataset that includes communication, visual, and LiDAR information. However, it does not cover UAV-related scenarios, which restricts its usefulness for low-altitude or aerial mobility research. DeepSense 6G \cite{alkhateeb2023deepsense} is a measurement-based multi-modal dataset that combines sensing and communication data. Nevertheless, it lacks depth maps, event camera outputs, and UAV scenarios. Furthermore, the high cost and limited flexibility of real-world data collection hinder its adaptability to customized multi-modal ISAC environments. In contrast, Great-MCD provides the largest-scale multi-modal ISAC dataset developed specifically for UAV scenarios.

\begin{figure*}[!t]
\centering\includegraphics[width=1\textwidth]{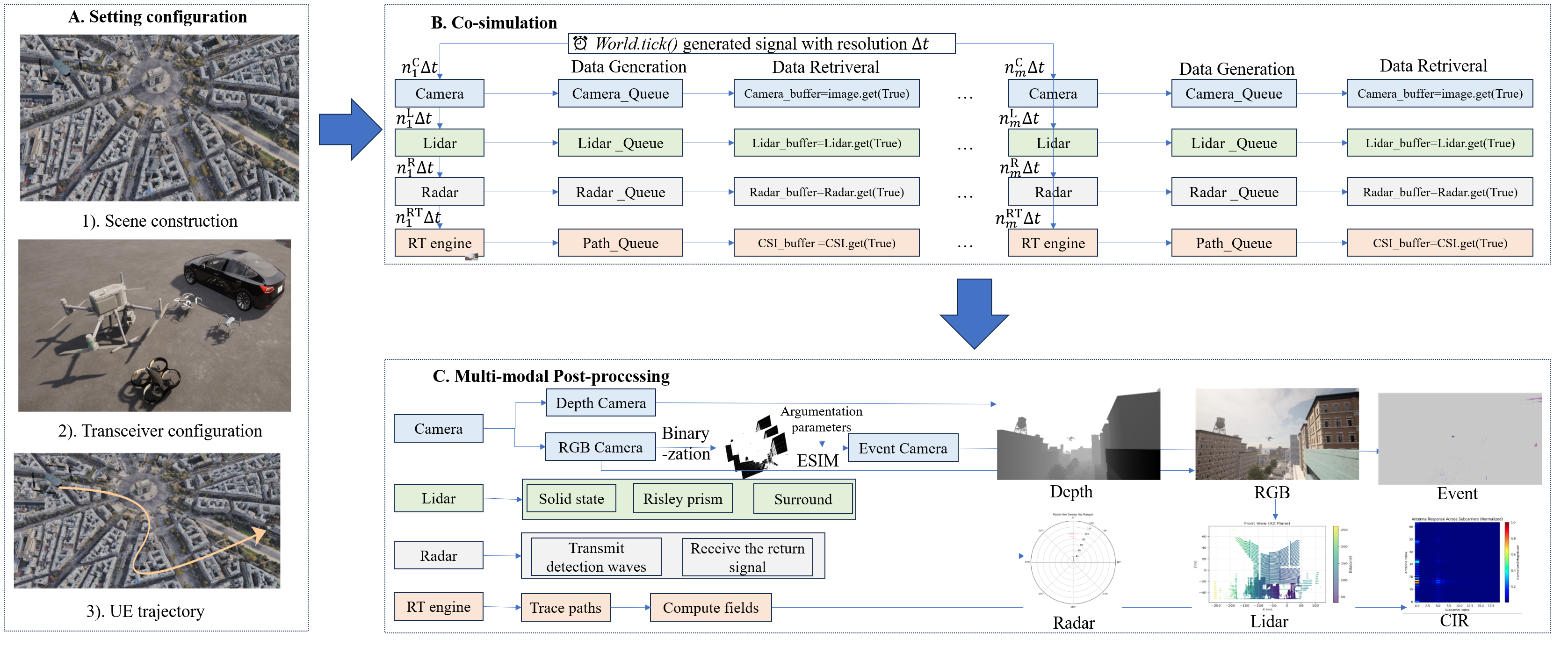}
\caption{The unified multi-modal co-simulation architecture of the Great-X framework, comprising three interconnected modules: \textbf{A. Setting Configuration}, \textbf{B. Core Co-Simulation}, and \textbf{C. Multi-modal Post-Processing}. The \textit{world.tick()} mechanism in Unreal Engine guarantees zero jitter and precise time alignment in multi-sensor timestamping for multi-modal data generation. }
\label{fig:Great-X_test}
\end{figure*}

\section{Pipeline of Great-X and Great-MCD}
\label{sec:greatx}

The overall architecture of the proposed Great-X framework is illustrated in Fig.~\ref{fig:Great-X_test}, which comprises three interconnected modules: setting configuration, core co-simulation, and multi-modal post-processing. Unlike conventional simulators that rely on fragmented software stacks, Great-X leverages the high-fidelity rendering and physics capabilities of Unreal Engine to provide a unified platform for ISAC research.

\subsection{Setting Configuration}
Before running any simulation, several parameters must be defined to produce consistent and realistic results. These cover scene construction with high-resolution mapping, transceiver settings, and user equipment (UE) trajectories.

\subsubsection{Scene Construction} Great-X provides flexible environmental modeling that differs from conventional simulators, as shown in Fig.~\ref{fig:Great-X_test}. The platform uses a dual-mode map generation pipeline. One mode performs automated three-dimensional reconstruction directly from high-resolution satellite imagery. The other incorporates detailed custom assets from the Unreal Engine ecosystem. This flexibility enables support the creation of accurate, large-scale scenes required for the research on 6G. 

\subsubsection{Transceiver Configuration} The transceiver module provides the resolution and flexibility needed for RF front-end and physical-layer modeling, ensuring compatibility with 6G channel requirements.
\begin{inparaenum}[\itshape a\upshape)]
\item The system supports ideal isotropic antennas, half-wave dipoles, and built-in three-dimensional radiation patterns that follow the 3GPP TR 38.901 standard. Users may freely set array size and element spacing for both uniform linear arrays (ULA) and uniform planar arrays (UPA).
\item In the polarization domain the module accommodates vertical polarization ($0^\circ$), horizontal polarization ($90^\circ$), dual polarization (VH), and cross polarization ($\pm 45^\circ$). A rotation matrix defined by the slant angle converts any ray into the orthogonal zenith ($\theta$) and azimuth ($\phi$) components, yielding the complex field quantities $E_\theta$ and $E_\phi$.
\item The transceiver is coupled to the engine's environmental variables, such as the atmospheric humidity, and to the kinematics module. It therefore computes distance- and material-dependent path loss on the fly and obtains the instantaneous three-dimensional velocity vector of every node for precise Doppler-shift calculation.
\item Finally, all antenna field data are converted at the lowest level into a standard multi-dimensional complex tensor. This compact format supplies physically grounded inputs to downstream multi-modal fusion networks and maintains interoperability with other channel simulators.
\end{inparaenum}
\subsubsection{UE Trajectory} 
Great-X includes a versatile trajectory generator that support continuous trajectory generation and discrete trajectory uploading. 
\begin{itemize}
\item Continuous trajectory generation: Traditional rigid waypoint schemes are replaced by adaptive strategies. Inside confined spaces, such as corridors, the generator applies a spatially constrained random linear interpolation model. For wide-area outdoor missions (for example UAV low-altitude patrols) it employs a four-point parametric Bézier curve in three dimensions. Users could define a bounding box with explicit limits on the $X$, $Y$, and $Z$ coordinates together with the minimum and maximum allowed distances ($d_{\min}$, $d_{\max}$) between transceiver nodes. Within these bounds the algorithm draws the start point $P_0$ and end point $P_3$ from an external random seed. Configurable spatial offsets are then injected at the midpoint to create the intermediate control points $P_1$ and $P_2$. The resulting smooth trajectory is obtained from the cubic Bézier equation:
\begin{equation}
B(t)=(1-t)^3P_0+3(1-t)^2tP_1+3(1-t)t^2P_2+t^3P_3,    
\end{equation}
which guarantees continuous velocity and acceleration in time interval $t\in[0,1]$.
\item Discrete trajectory uploading: The same generator also accepts discrete time-coordinate sequences recorded in real-world experiments, permitting exact one-to-one reproduction of measured trajectories inside the simulator.
\end{itemize}

Regardless of whether a trajectory is generated parametrically or imported from measurements, the UE’s spatial displacement remains locked to the engine’s fixed time step (such as $\Delta t=0.001$\,s). This frame-by-frame advancement supplies accurate instantaneous velocities ($v=\Delta p/\Delta t$) for Doppler modeling and guarantees exact spatiotemporal alignment among CSI, RGB images, depth maps, and LiDAR point clouds at every sampling instant.

\subsection{Co-Simulation}

Fig.~\ref{fig:Great-X_test} shows the core co-simulation and synchronization scheduling module, which functions as the central operating engine of the Great-X framework. In multi-modal 6G research, sensors that run at different sampling rates must still deliver data with exact spatiotemporal alignment. This requirement is one of the principal obstacles to building high-fidelity datasets. Great-X meets the requirement by enforcing a \textit{deterministic synchronous mode} at the lowest level of the engine.

Under this mode a single global reference clock controls every component. The physics engine advances the simulation in fixed time interval (such as $\Delta t = 0.001$\,s), which serve as the highest available temporal resolution. Although the physical capture cycles of the various sensors differ,e.g., visual cameras at 30\,Hz, LiDAR at 10\,Hz, RF transceivers at arbitrary frequencies, their trigger instants are locked to integer multiples of $\Delta t$. Consequently, simulation time is driven exclusively by the engine's \textit{world.tick()} signal and remains independent of wall-clock time on the host computer.

Inside each simulation frame, the system executes a deterministic sequence of operations. It first updates the absolute three-dimensional coordinates of the UE for the current timestamp. Every sensor mounted on the vehicle is then repositioned according to its fixed 6-DoF rigid-body offset. Data produced by the active sensors are written in parallel to separate thread-safe queues. A central buffer manager withholds the current frame until all required modalities, including the electromagnetic ray-tracing results, have arrived and share an identical timestamp. Only at that point does the manager release the synchronized data packet.

This design preserves the native, heterogeneous sampling behavior of each sensor while completely removing motion blur and spatial misalignment. The result is strict frame-level temporal alignment with zero drift among all data streams at every sampling instant.

\begin{figure*}[!t]
    \centering
\subfigure[Distribution of Great-MCD.]{
\includegraphics[width=.9\columnwidth]{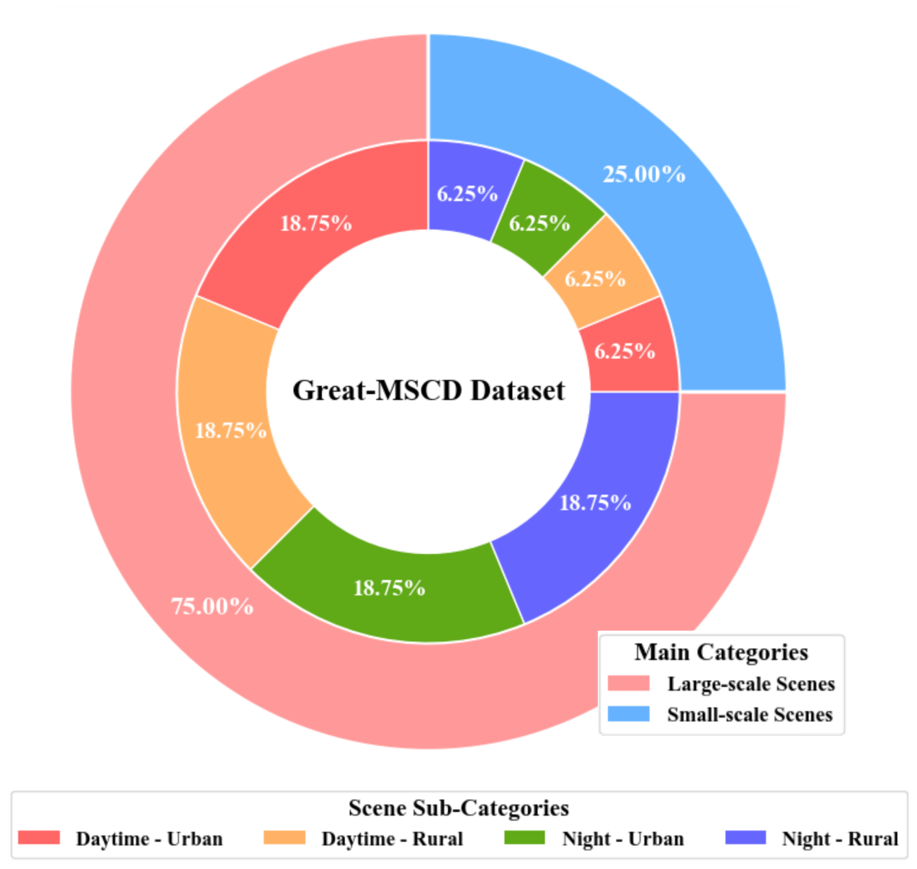}\label{fig:data_dis}
}
\subfigure[Configurations of Great-MCD.]{
\includegraphics[width=1.03\columnwidth]{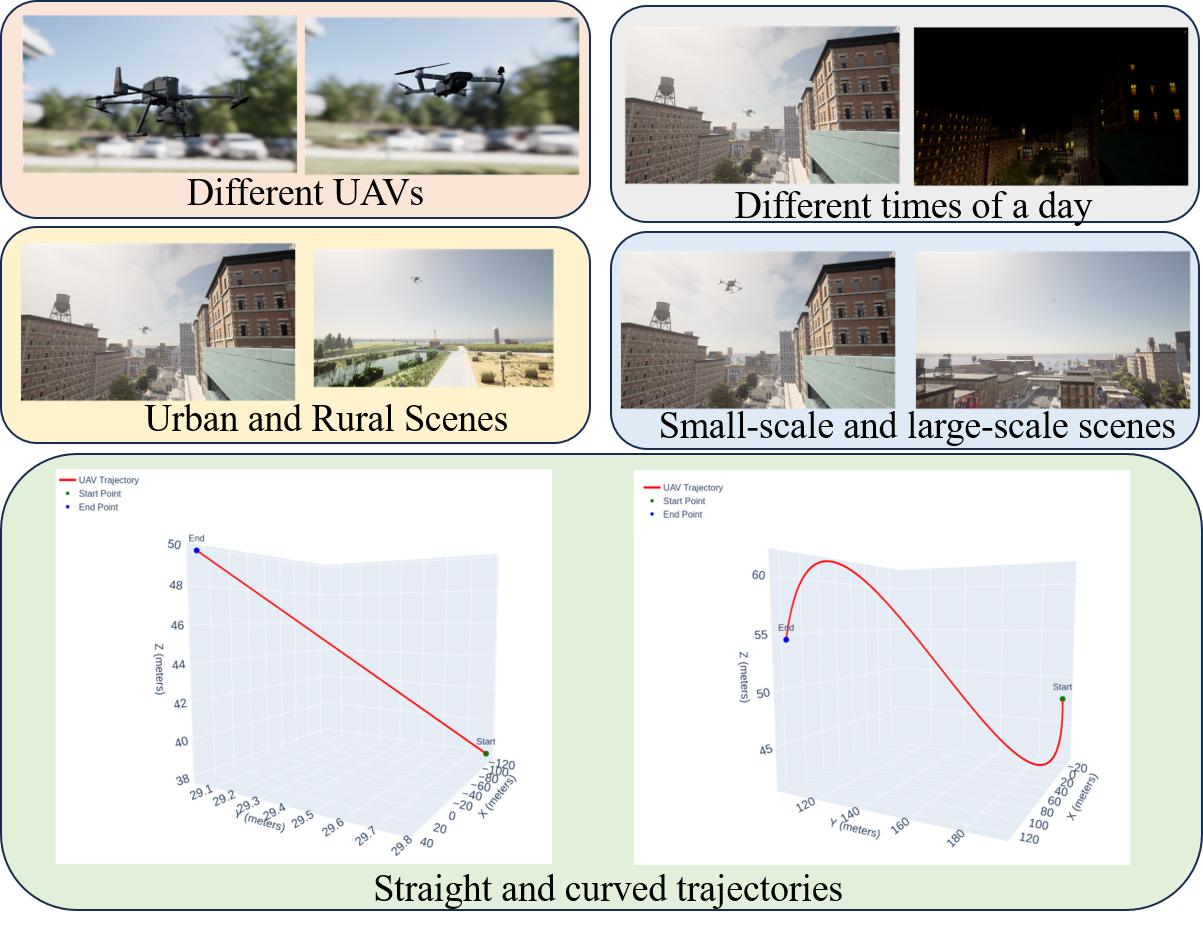}\label{fig:data_config}
}\caption{Configuration of GREAT-MCD, including day/night, UAV types, different scene sizes, trajectory types and rural/urban scenarios. The detail simulation parameters can be found in Table \ref{tab:Combined_Config}.}
\label{fig:great_MSCD}
\end{figure*}

\subsection{Multi-modal Post-Processing}
Fig.~\ref{fig:Great-X_test} illustrates the multi-modal post-processing module, which converts raw outputs from the physics engine into data formats suitable for AI-driven 6G research. Raw engine data, such as uncalibrated depth buffers and discrete ray-tracing paths, cannot be used directly for neural-network training. Great-X therefore applies a standardized pipeline that converts heterogeneous sensor outputs into modality-specific feature representations tailored to 6G tasks:
\begin{itemize}
\item Visual, Neuromorphic, and Spatial Geometric Reconstruction:
For optical sensors the pipeline produces normalized RGB images and reconstructs depth information from the raw 24-bit buffer. A logarithmic transformation $\log(x+1)$ is applied to expand the dynamic range of distant features. In addition, Great-X incorporates an event-based vision branch that emulates the asynchronous behavior of event cameras. Using conversion logic derived from event-based vision simulator (ESIM), the system computes intensity differences across consecutive RGB frames, performs differential binarization, and applies parametric augmentation to generate asynchronous event streams with high temporal resolution. This modality is particularly useful for capturing fast transients during high-speed UAV maneuvers. For LiDAR, the pipeline directly extracts a four-dimensional matrix $(x, y, z, I)$ of spatial coordinates and reflection intensity from the engine memory, preserving the full geometric information without loss.
\item Electromagnetic Feature Domain Transformation and High-Fidelity CSI: 

In the RF domain, the post-processing module reconstructs the electromagnetic propagation that occurs inside Unreal Engine. The algorithm first separates line-of-sight (LOS) and non-line-of-sight (NLOS) components and aggregates the spatial angles ($\theta, \phi$) together with the time delays of all reflection paths. Linear amplitude attenuation factors that depend on atmospheric humidity are then superimposed to convert the discrete ray parameters into a continuous channel impulse response (CIR). Finally, a fast Fourier transform (FFT) maps the CIR into the frequency domain, yielding the frequency-domain channel state information $H(f)$ that retains both physical interpretability and high fidelity.

\item Framework-level Restructuring and Ground-Truth Encapsulation: 
To enable seamless operation within modern AI workflows, the original TensorFlow-based Sionna channel modeling module was refactored at a low level and its core logic was migrated to the NumPy and PyTorch ecosystems. As a result, all frequency-domain CSI data are encapsulated as native PyTorch complex tensors whose dimensions are precisely aligned: [Batch, No of Rx, $Rx_\text{ant}$, No of Tx, $Tx_\text{ant}$, $Time_\text{steps}$, Subcarriers]. At the same time the system extracts kinematic ground truth—including absolute coordinates and timestamps—and writes it to disk together with the corresponding images (.png), point clouds, and CSI matrices (.npy) using matching frame indices. This arrangement supplies a ready-to-use digital-twin dataset for learning-based communications and joint ISAC optimization tasks.
\end{itemize}

\begin{figure*}[!t]
\centering\includegraphics[width=0.7\textwidth]{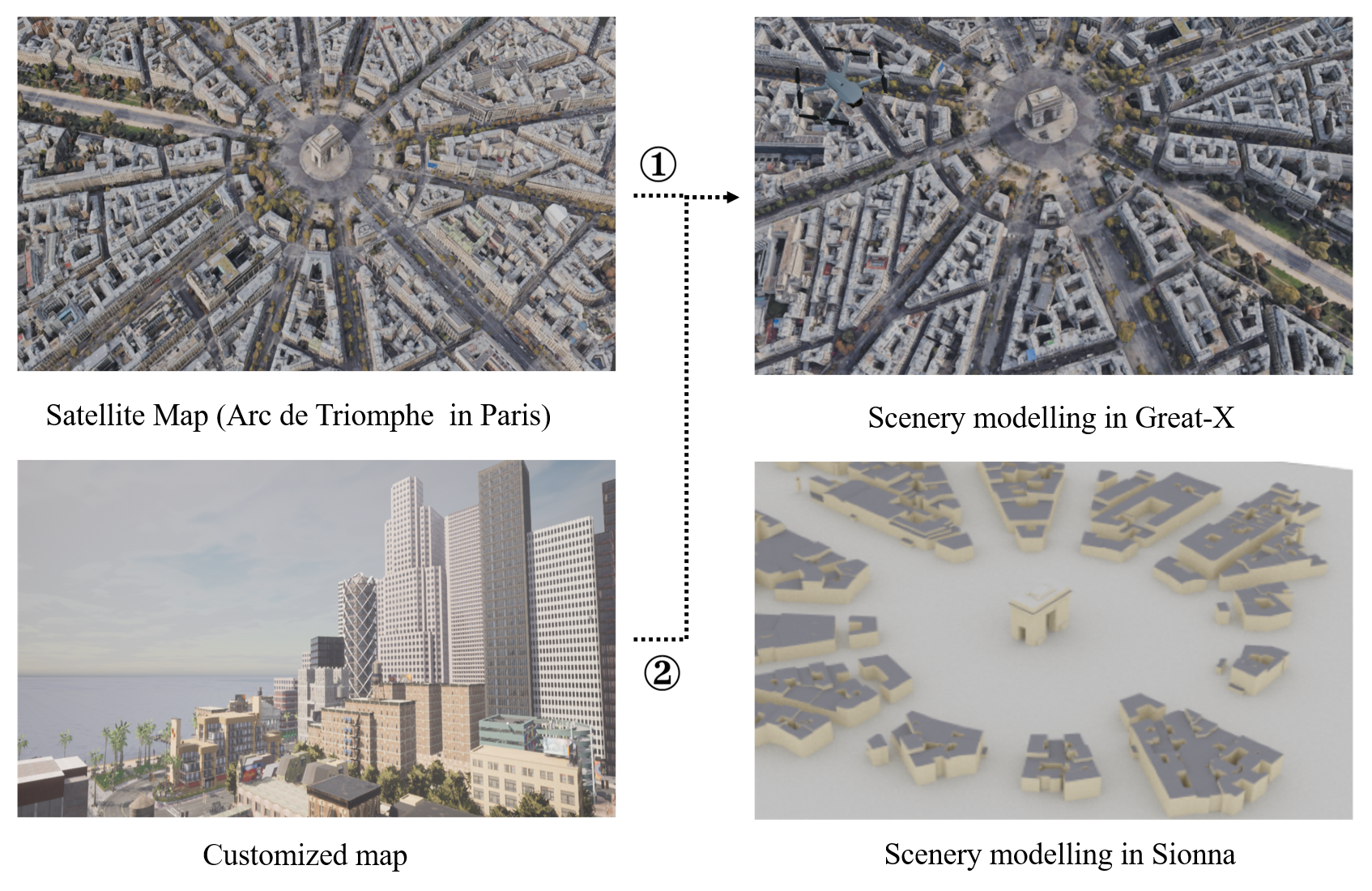}
\caption{The comparison between the geometric and semantic fidelity of Great-X and Sionna. An exemplified Arc de Triomphe scenario can be generated by Great X via either high-resolution satellite imagery or sophisticated custom assets from the Unreal Engine ecosystem. }
\label{fig:Arc _de_Triomphe}
\end{figure*}

\begin{table}[tbp]
\centering
\caption{Key Configuration Parameters of the Great-MCD Dataset.}
\label{tab:Combined_Config}
\resizebox{\columnwidth}{!}{%
\begin{tabular}{ll}
\hline
{Parameter}                                 & {Value}                                 \\ \hline

{Time Modes}                                & Day / Night                                    \\
{UAV Types}                                 & Large, Medium, Small                           \\
{Camera Field of View ($^\circ$)}              &  $90$                                            \\
{Camera Rotation ($^\circ$)}                       & {[}10, 170, 0{]}                               \\
{BS Rotation ($^\circ$)}                           & {[}0, 0, 0{]}                                  \\
{Transmitting Antenna Array}                & $4\times8$ UPA, Cross-polarized                       \\
{Antenna Plane (Tx)}                        & yz-plane                                       \\
{Vertical Spacing (Tx)}                     & $2 \lambda$                                      \\
{Horizontal Spacing (Tx)}                   & $0.5 \lambda$                                      \\
{Receiving Antenna Structure}               & $1\times1 $                                           \\ \hline
\end{tabular}%
}
\end{table}
\subsection{GREAT-MCD}
Following the simulation pipeline described above, we constructed the Great-MCD dataset. The dataset consists of long-duration dynamic scenes that exhibit strong temporal continuity and sequential dependencies. In a typical scenario, a UAV takes off from a distant location and follows a predefined trajectory toward a base station. Throughout the flight we simultaneously record RGB images, depth maps, position data, and CSI. These synchronized measurements capture the evolution of the wireless channel as the UAV moves through space, providing a useful resource for downstream tasks such as channel modeling, object detection, and location prediction.

As illustrated in Fig.~\ref{fig:great_MSCD}, the dataset contains two primary environment types: urban and rural. Each type is further divided into small-scale and large-scale sub-scenes that correspond to different UAV mission profiles. In urban areas the small-scale sub-scenes represent low-altitude flights among buildings, whereas the large-scale sub-scenes depict higher-altitude flights above rooftops. In rural areas the small-scale sub-scenes correspond to flights along rural roads, while the large-scale sub-scenes cover wide-area operations over open farmland.

For the small-scale scenes in both urban and rural settings we define a three-dimensional volume measuring 170\,m in length, 16\,m in width, and 30\,m in height. Within each volume we randomly generate 100 straight flight trajectories, each sampled at 200 uniformly spaced points. For the large-scale scenes we define a larger volume of 170\,m $\times$ 114\,m $\times$ 60\,m and generate 200 random curved trajectories, each containing 300 uniformly sampled data points.

To enhance dataset diversity and improve generalization, every sub-scene is rendered under both daytime and nighttime lighting conditions. We also include multiple UAV models and scenarios in which vehicles and drones interact simultaneously with the base station. This combination of varied environments, lighting conditions, vehicle types, and interaction patterns creates a rich testbed for multi-modal learning, environmental perception, and systematic evaluation of model generalization in complex settings. The key configuration parameters of the simulation setup are summarized in Table~\ref{tab:Combined_Config} for clarity and reproducibility.

\begin{figure*}[!t]
    \centering
\subfigure[TM polarization.]{
\includegraphics[width=0.75\columnwidth]{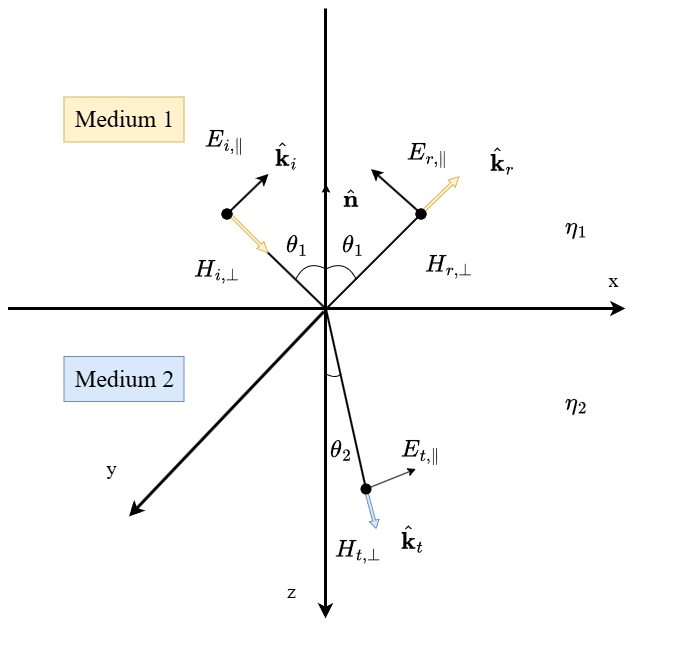}\label{fig:TM
}}
\subfigure[TE polarization.]{
\includegraphics[width=0.75\columnwidth]{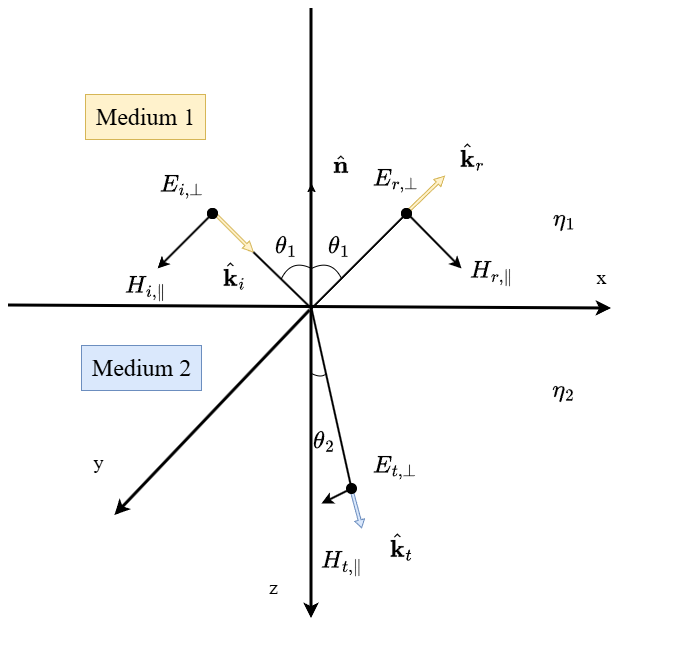}\label{fig:TE}
}\caption{Interaction of a plane electromagnetic wave with a planar boundary between two media, involving reflection and transmission.}
    \label{fig:reflection_refraction}
\end{figure*}

\section{Core Modules and Key Features of Great-X}
\label{sec:core_modules}
\subsection{Unified High-Fidelity Scene Construction}
Fig.~\ref{fig:Arc _de_Triomphe} compares the geometric and semantic fidelity of Great-X with that of baseline simulators, e.g., Sionna. In the Arc de Triomphe scenario, for example, Great-X reconstructs fine environmental details such as precise building facets, varied vegetation, and realistic road textures obtained from satellite imagery. By comparison, simulators such as Sionna, which rely on the Mitsuba renderer, typically employ simplified geometric primitives and therefore cannot capture the spatial granularity of complex urban foliage or intricate structural outlines.

These differences in scene representation directly affect the reliability of the simulated data. The coarse geometry used by Sionna often produces oversimplified ray-tracing paths and tends to overestimate sensing performance. In contrast, the higher level of detail provided by Great-X enables the generated modalities, particularly CSI multipath components and LiDAR point clouds, to accurately reproduce the scattering and occlusion effects present in real electromagnetic environments.

\subsection{Native Electromagnetic Modeling}

Great-X implements a physically accurate far-field electromagnetic model directly inside Unreal Engine, enabling high-fidelity CSI generation that remains perfectly synchronized with visual and geometric modalities. The far-field electric field of the transmitting antenna with respect to the radial distance from the antenna $r$, zenith angle $\theta$, and azimuth angle $\varphi$ is given as
\begin{equation}
\mathbf{E}_{\mathrm{T}}(r,\theta_{\mathrm{T}},\varphi_{\mathrm{T}}) =
\sqrt{\frac{P_{\mathrm{T}} G_{\mathrm{T}} Z_{0}}{2\pi}}
\frac{e^{-jk_{0}r}}{r}
\mathbf{F}_{\mathrm{T}}(\theta_{\mathrm{T}},\varphi_{\mathrm{T}}),
\end{equation}
where $P_{\mathrm{T}}$ is the input power of the transceiver. $G_{\mathrm{T}}$ and $\mathbf{F}_{\mathrm{T}}(\theta_{\mathrm{T}},\varphi_{\mathrm{T}})$ are the antenna gain of the transceiver and the normalized complex antenna radiation pattern, respectively, that incorporate realistic 3GPP-compliant radiation patterns and polarization. $Z_0$ is the impedance of free space.

At the receiver, the incoming wave is treated as planar. The effective aperture and open-circuit voltage are derived while accounting for arbitrary polarization and direction of arrival:
\begin{equation}
V_{\mathrm{R}} = \sqrt{\frac{\lambda^{2}}{4\pi}G_{\mathrm{R}}\frac{4\Re\left\{Z_{\mathrm{R}}\right\}}{Z_{0}}}\left\|\mathbf{E}_{\mathrm{T}}(r,\theta_{\mathrm{T}},\varphi_{\mathrm{T}})\right\|^{2},
\end{equation}where $G_{\mathrm{R}}$ and $Z_\text{R}$ are the antenna gain and impedance of the receiver, respectively.

Because this electromagnetic model is natively coupled with the engine's high-fidelity 3D geometry, material properties, and deterministic synchronous clock, Great-X produces channel impulse responses and CSI that closely match real-world measurements while guaranteeing zero-drift alignment with RGB, depth, and LiDAR data. This unified physical modeling is a key reason for the simulator’s strong Sim2Real transfer performance in 6G tasks.

\subsection{Zero-Drift Multi-Modal Synchronization}
Co-simulating heterogeneous sensors across separate software platforms commonly suffers from latency jitter and clock drift, which seriously degrades the temporal alignment of multi-modal datasets. Although a single-engine architecture can eliminate inter-process communication delays, it does not automatically guarantee synchronization. Rendering pipelines and physics computations in game engines typically run at variable rates that depend on instantaneous computational load.

Great-X overcomes this limitation through a deterministic synchronization strategy built on a strict clock model and spatially aligned coordinates. Unlike conventional engines that employ variable time steps tied to real-time rendering demands, Great-X enforces a fixed-time-step clock. The entire simulation advances solely by a constant, predefined interval $\Delta t$ at each engine tick. Simulation time is therefore completely decoupled from wall-clock time on the host computer and progresses in a fully deterministic manner.

This fixed-step mechanism locks the trigger instants of all sensors,visual cameras, LiDAR, RF transceivers, and event cameras,to integer multiples of the base interval. As a result, Great-X achieves strict frame-level temporal alignment with zero drift across CSI, RGB images, depth maps, and LiDAR point clouds at every sampling instant. Such rigorous multi-modal synchronization is a key advantage of the unified single-engine design and constitutes an essential foundation for generating high-quality data that transfers effectively from simulation to real-world 6G ISAC systems.

\subsection{Rigorous Ray-Tracing for CSI}
Great-X reconstructs electromagnetic wave propagation directly inside Unreal Engine through native ray tracing, enabling physically accurate, ray-based channel modeling that is tightly coupled with the engine’s high-fidelity 3D geometry and materials. This unified implementation forms a core advantage over conventional multi-platform simulators. Following standard ray-tracing principles, interactions are computed using signal path decomposition and material response models. We focus here on the reflection component, which dominates multipath behavior in typical environments.

When a plane wave encounters a planar interface between two media, part of the energy is reflected while the remainder is transmitted. For clarity, we assume uniform, non-magnetic dielectrics and adopt material parameters recommended by the ITU. As illustrated in Fig.~\ref{fig:reflection_refraction}, the incident field is decomposed into transverse-electric (TE) and transverse-magnetic (TM) polarization components. The corresponding reflected and transmitted fields are obtained via the Fresnel reflection and transmission coefficients, which depend on the incidence angle, the relative permittivities of the two media, and the polarization state.

The wave vectors of the reflected and transmitted rays are determined by the law of reflection and Snell’s law, respectively. Combining these relations yields the following expressions for the reflected and transmitted field components:
\begin{subequations}
\begin{align}
\begin{bmatrix}
E_{\mathrm{r},\perp} \\
E_{\mathrm{r},\parallel}
\end{bmatrix}
&=
\begin{bmatrix}
r_{\perp} & 0 \\
0 & r_{\parallel}
\end{bmatrix}
\mathbf{W}
\begin{bmatrix}
E_{\mathrm{i},s} \\
E_{\mathrm{i},p}
\end{bmatrix}, \\
\begin{bmatrix}
E_{\mathrm{t},\perp} \\
E_{\mathrm{t},\parallel}
\end{bmatrix}
&=
\begin{bmatrix}
t_{\perp} & 0 \\
0 & t_{\parallel}
\end{bmatrix}
\mathbf{W}
\begin{bmatrix}
E_{\mathrm{i},s} \\
E_{\mathrm{i},p}
\end{bmatrix},
\end{align}
\label{eq:reflection_transmission}
\end{subequations}
where $r_{\perp}$, $r_{\parallel}$, $t_{\perp}$, and $t_{\parallel}$ are the Fresnel coefficients, and $\mathbf{W}$ is the polarization transformation matrix. When the incidence angle exceeds the critical angle, total internal reflection occurs and the transmission coefficients become zero.

These electromagnetic interactions were implemented natively in Unreal Engine through C++ refactoring. Because the ray-tracing engine shares the same high-fidelity 3D meshes and physically based rendering materials used for visual rendering, Great-X produces channel impulse responses and CSI that accurately capture real-world scattering and occlusion effects while remaining perfectly synchronized with RGB, depth, and LiDAR data. This tight integration of rigorous electromagnetic modeling with the unified simulation engine is a primary reason for the simulator’s strong physical fidelity and effective Sim2Real transfer performance.

\subsection{High-Speed Event-Based Perception}
To support high-speed perception in dynamic UAV scenarios, Great-X incorporates an event-camera modality that is rarely available in existing ISAC simulators. We emulate the asynchronous brightness-sensing behavior of real event cameras using conventional RGB sequences inside the unified simulation engine. Given a sequence of RGB frames $\{I_t\}$ captured at a fixed frame rate $f$, the framework produces both per-frame event images and a continuous asynchronous event stream. This approach bridges standard visual sensing with event-based sensing while maintaining perfect temporal alignment with other modalities such as CSI, depth, and LiDAR.

For consecutive frames $I_{t-1}$ and $I_t$, the gray-scale intensity difference is first computed as
\begin{equation}
D_t(x,y) = I_t(x,y) - I_{t-1}(x,y),
\end{equation}
where $(x,y)$ denotes pixel coordinates. Pixels whose intensity change exceeds a positive threshold $\tau_p$ are labeled as positive events, while those whose change falls below a negative threshold $-\tau_n$ are labeled as negative events:
\begin{equation}
E_t(x,y) =
\begin{cases}
+1, & D_t(x,y) > \tau_p, \\
-1, & D_t(x,y) < -\tau_n, \\
0, & \text{otherwise.}
\end{cases}
\end{equation}
The resulting event image visualizes intensity transitions across frames, with static regions shown in neutral gray to highlight dynamic changes.

To better approximate the behavior of commercial event cameras such as dynamic and active-pixel vision sensor (DAVIS) or Prophesee, we additionally employ a logarithmic contrast model. For each pixel, the change in logarithmic brightness is computed as
\begin{equation}
\Delta L_t(x,y) = \log\!\left(I_t(x,y) + \epsilon\right) - \log\!\left(I_{t-1}(x,y) + \epsilon\right),
\end{equation}
where $\epsilon$ is a small positive constant that prevents division by zero. An event is generated whenever this change exceeds a predefined contrast threshold $C$. Each event is stored as the tuple $(x, y, t, p)$, where the polarity $p \in \{+1, -1\}$ indicates whether brightness increased or decreased, and the timestamp $t = i/f$ is derived from the frame index $i$ and frame rate $f$. The complete event stream is therefore expressed as
\begin{equation}
\mathcal{E} = \{(x_k, y_k, t_k, p_k)\}_{k=1}^{N},
\end{equation}
where $N$ is the total number of generated events.

As this event simulation runs natively within the same deterministic clock and unified 3D engine as all other sensors, the generated event streams maintain exact spatiotemporal alignment with CSI, RGB, depth, and LiDAR data. This capability provides a distinctive advantage for modeling transient phenomena during high-speed UAV maneuvers and enriches the multi-modal dataset for downstream 6G ISAC tasks.

\section{Simulation results and analysis}
To provide a comprehensive evaluation of the proposed Great-X simulator, we first carry out validation with the measurement dateset, then come two distinct tasks. The first is a primary multi-modal 3D positioning task designed to assess cross-modal fusion capabilities, while the second is a downstream CSI feedback task intended to verify the physical-layer fidelity of the simulated wireless channels.

To ensure a rigorous and fair comparison, all cross-platform evaluations are conducted within a uniform simulation environment. By utilizing an identical map across all benchmarks, we isolate the architectural and algorithmic contributions of Great-X from the influence of heterogeneous environmental modeling. It should be noted, however, that while this controlled setup ensures parity in the evaluation process, it may not fully reflect the superior modeling granularity of Great-X as illustrated in Fig.~\ref{fig:Arc _de_Triomphe}. Consequently, the quantitative results presented in this section represent a conservative estimate of the performance gains achievable with the full modeling potential of our framework.


\label{sec:experiments}
\begin{figure}[!t]
  \centering 
    \centering
    \includegraphics[width=\linewidth]{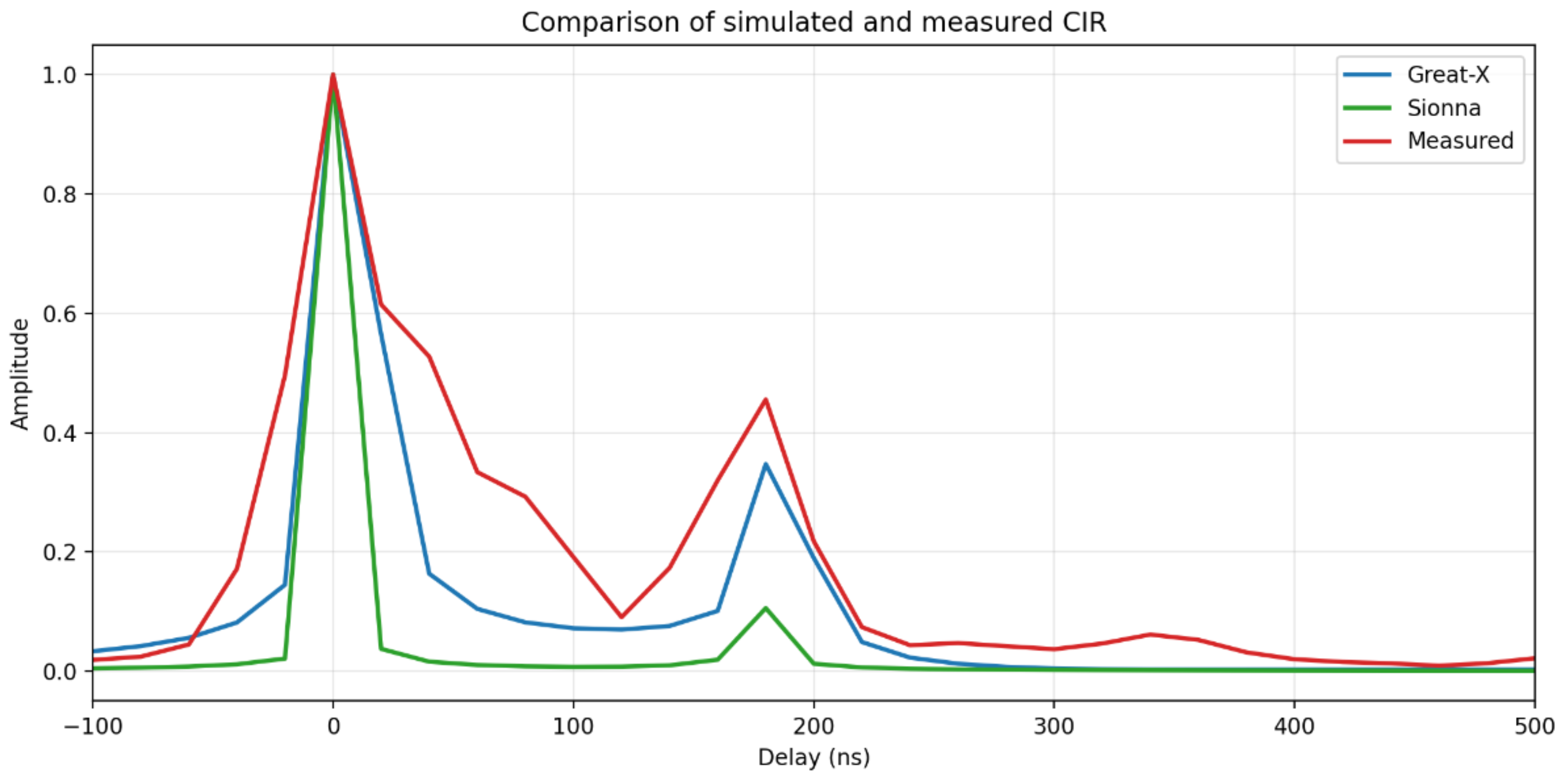}
    \caption{ Channel impulse response (CIR) comparison among Great-X, Sionna, and measured results.}
    \label{fig:CIR_compare}
\end{figure}

\subsection{Simulator Validation: Comparison of Simulated and Measured Data}
Before deploying simulated data for downstream tasks, it is crucial to rigorously validate the physical fidelity of the Great-X simulator. To this end, we conduct a comparative study using the well-established real-world DICHASUS measurement dataset collected by the Institute of Telecommunications at the University of Stuttgart \cite{dataset-dichasus-dcxx}. 

Real-world measurement environments were conducted in a representative, complex indoor industrial/office environment \cite{dataset-dichasus-dcxx}. The physical layout features interconnected corridors, an open-plan laboratory area, and structural obstacles including concrete pillars, glass partitions, and metallic cabinets. The Tx array, consisting of distributed massive MIMO antennas, was deployed across the ceiling to provide widespread coverage. The Rx was mounted on a moving robotic platform, traversing complex trajectories that naturally form alternating LoS and NLoS conditions. The rich scattering nature of this environment, combined with the diverse material composition, presents a highly challenging testbed for validating the accuracy of ray-tracing-based channel models.

The experimental dataset was acquired using a distributed MIMO channel sounder operating at a carrier frequency of 3.438 GHz with a 50 MHz bandwidth. To faithfully replicate the measurement environment, we integrated the provided high-precision 3D architectural layout and material specifications into the Great-X platform, thereby constructing a digital twin of the experimental site. The fidelity of the simulator was subsequently validated by benchmarking its outputs against the empirical measured data.


As shown in the Fig.~\ref{fig:CIR_compare}, we observe that the CIR generated by Great-X closely matches the measured results in overall structure. In particular, the simulated and measured waveforms exhibit a highly consistent amplitude variation pattern over time, with the main difference appearing in the absolute power levels of certain multipath components. This strong alignment indicates that the ray-tracing core of Great-X effectively captures the dominant propagation mechanisms present in real environments.

These amplitude scaling differences can be attributed to the inherent complexity of real electromagnetic environments. The relative permittivity $\varepsilon_r$ and conductivity $\sigma$ of building materials are neither spatially uniform nor constant across frequencies. Instead, they may vary with location and exhibit dispersion characteristics, whereas the current simulation model represents them using simplified constant values. This simplification can introduce deviations in path loss estimation.

 \begin{table}
    \centering
    \caption{Performance Comparison of Positioning Error (in meters). The proposed Multi-modal Knowledge Distillation (MMKD) achieves the best performance across all metrics.}
    \label{tab:kfold_summary}
    \resizebox{\linewidth}{!}{ 
    \begin{tabular}{lcc}
    \toprule
    \textbf{Dataset / Method} & \textbf{Mean Error (m) $\pm$ Std} & \textbf{Median Error (m)} \\ \midrule
    Great-X (Baseline) & 0.3529 $\pm$ 0.0154 & 0.2698 \\
    Sionna & 0.3565 $\pm$ 0.0160 & 0.2872 \\
    QD (Quadriga) & 0.4743 $\pm$ 0.0155 & 0.3784\\
    Measured & 0.1840 $\pm$ 0.0101  & 0.1496 \\
    Great-X (MMKD) & 0.1598 $\pm$ 0.0079 & 0.1222\\ \midrule
    \end{tabular}
    } 
  \end{table}
\begin{figure}[!t]
  \centering 
    \includegraphics[width=\linewidth]{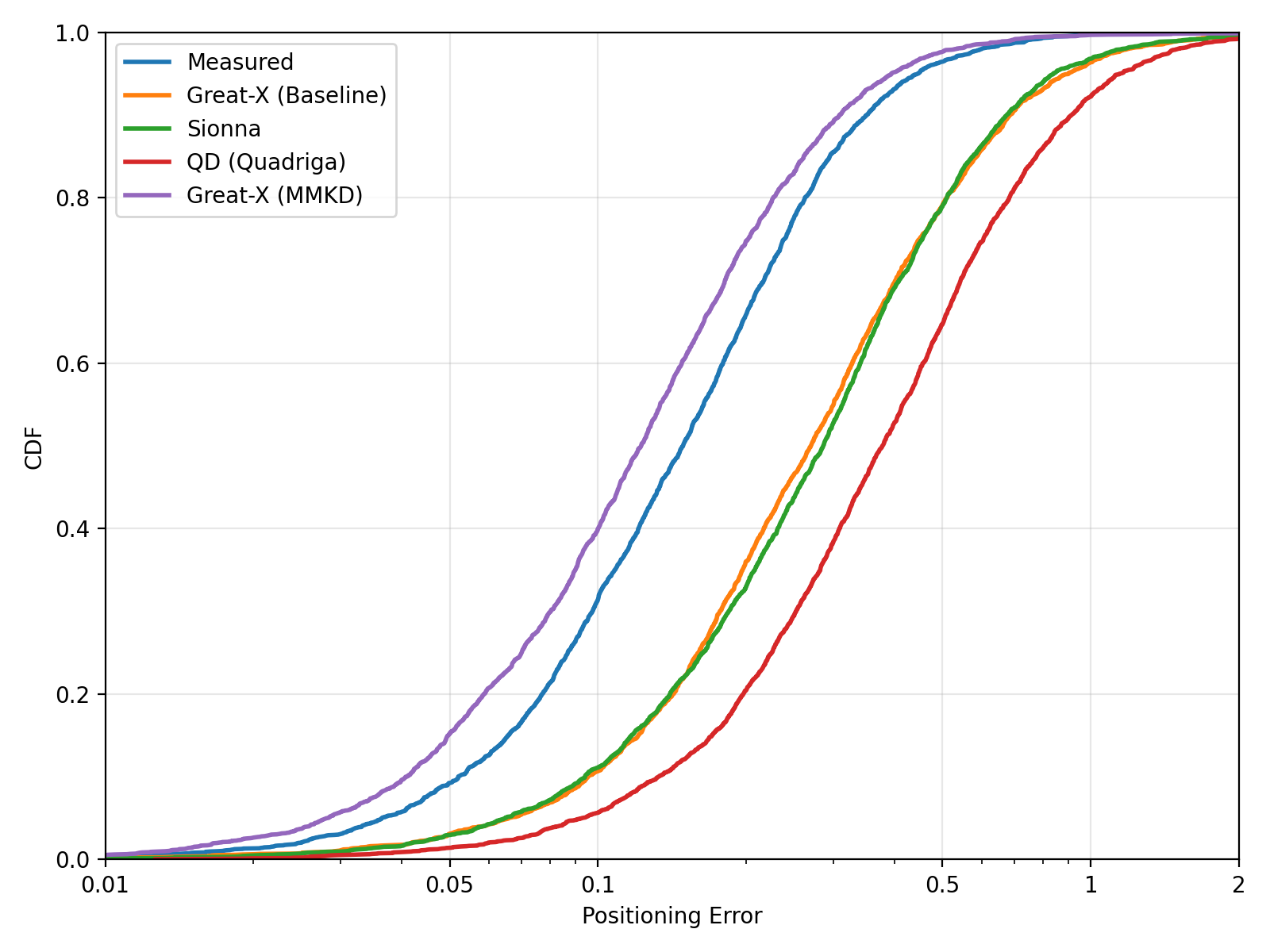}
    \caption{Cumulative Distribution Function (CDF) of positioning errors.}
    \label{fig:cdf_kfold}
\end{figure}

 \begin{table*}[htbp]
  \centering
    \centering
    \caption{Comparison of CSI Feedback Performance (NMSE in dB) across Different Simulators and Compression Ratios on Measured Data.}
    \label{tab:csi_results}
      \begin{tabular}{llcccc}
      \toprule
      \textbf{Dataset} & \textbf{Evaluation Stage} & \textbf{CR=1/4} & \textbf{CR=1/8} & \textbf{CR=1/16} & \textbf{CR=1/32} \\ \midrule
      \multirow{2}{*}{\textbf{Great-X}} & Stage 1 (Zero-shot) & 2.17 & 1.90 & 1.73 & 2.10 \\
       & Stage 2 (Fine-tuned) & -15.16 & -14.94 & -14.72 & -13.47 \\ \midrule
      \multirow{2}{*}{\textbf{Sionna}} & Stage 1 (Zero-shot) & 2.87 & 2.27 & 2.30 & 2.38 \\
       & Stage 2 (Fine-tuned) & -14.78 & -14.41 & -14.26 & -13.63 \\ \midrule
      \multirow{2}{*}{\textbf{QD}} & Stage 1 (Zero-shot) & 2.28 & 1.67 & 1.95 & 1.93 \\
       & Stage 2 (Fine-tuned) & -14.54 & -13.95 & -13.94 & -12.76 \\ \midrule
      \textbf{Measured Only} & Stage 2 (Direct-train) & -14.93 & -14.84 & -14.59 & -12.91 \\ \bottomrule
      \end{tabular}
\end{table*}

\begin{figure*}[!t]
    \centering
\subfigure[Compression ratio $1/4$.]{
\includegraphics[width=0.96\columnwidth]{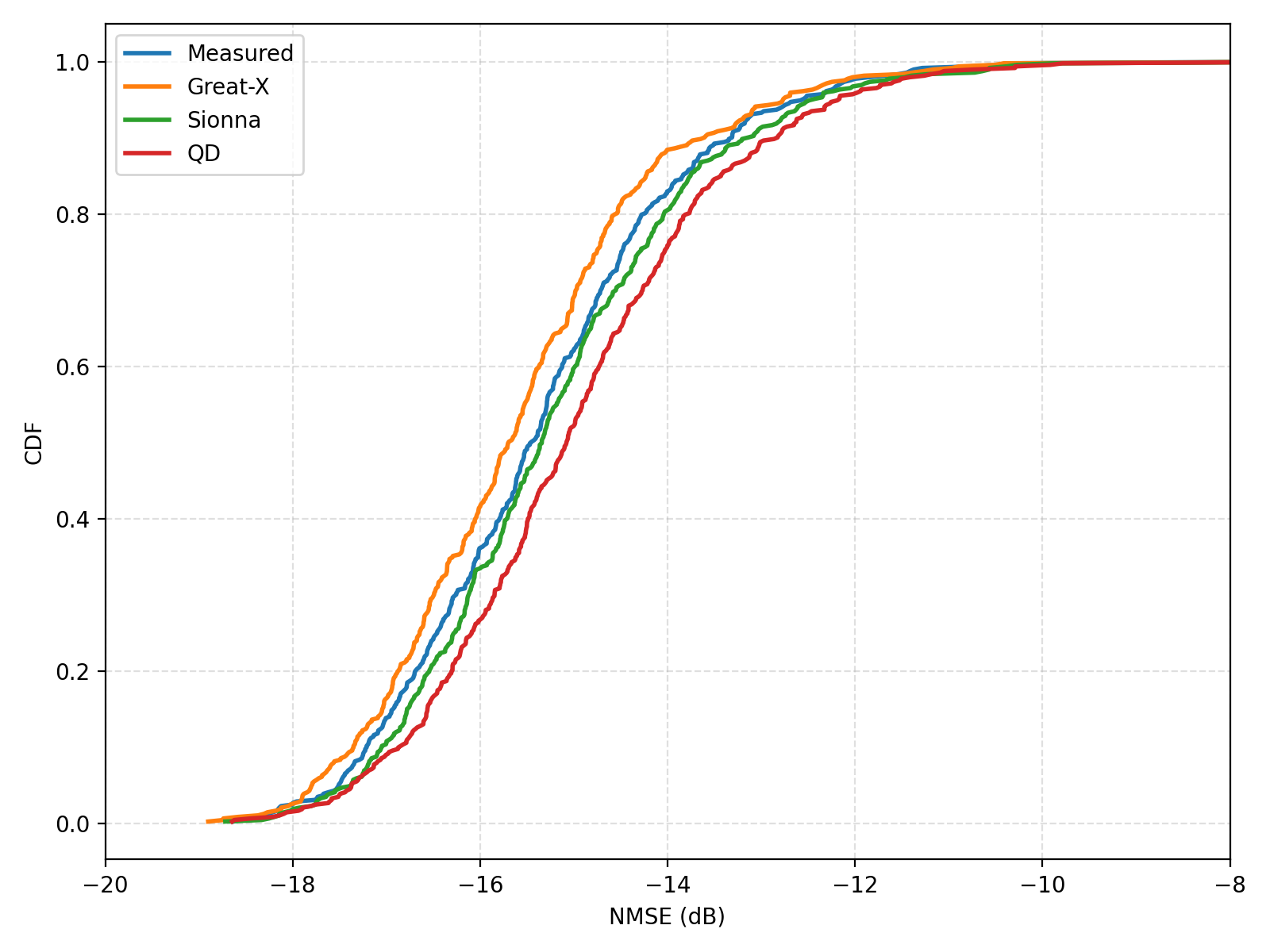}\label{fig:1_4}
}
\subfigure[Compression ratio $1/8$.]{
\includegraphics[width=0.96\columnwidth]{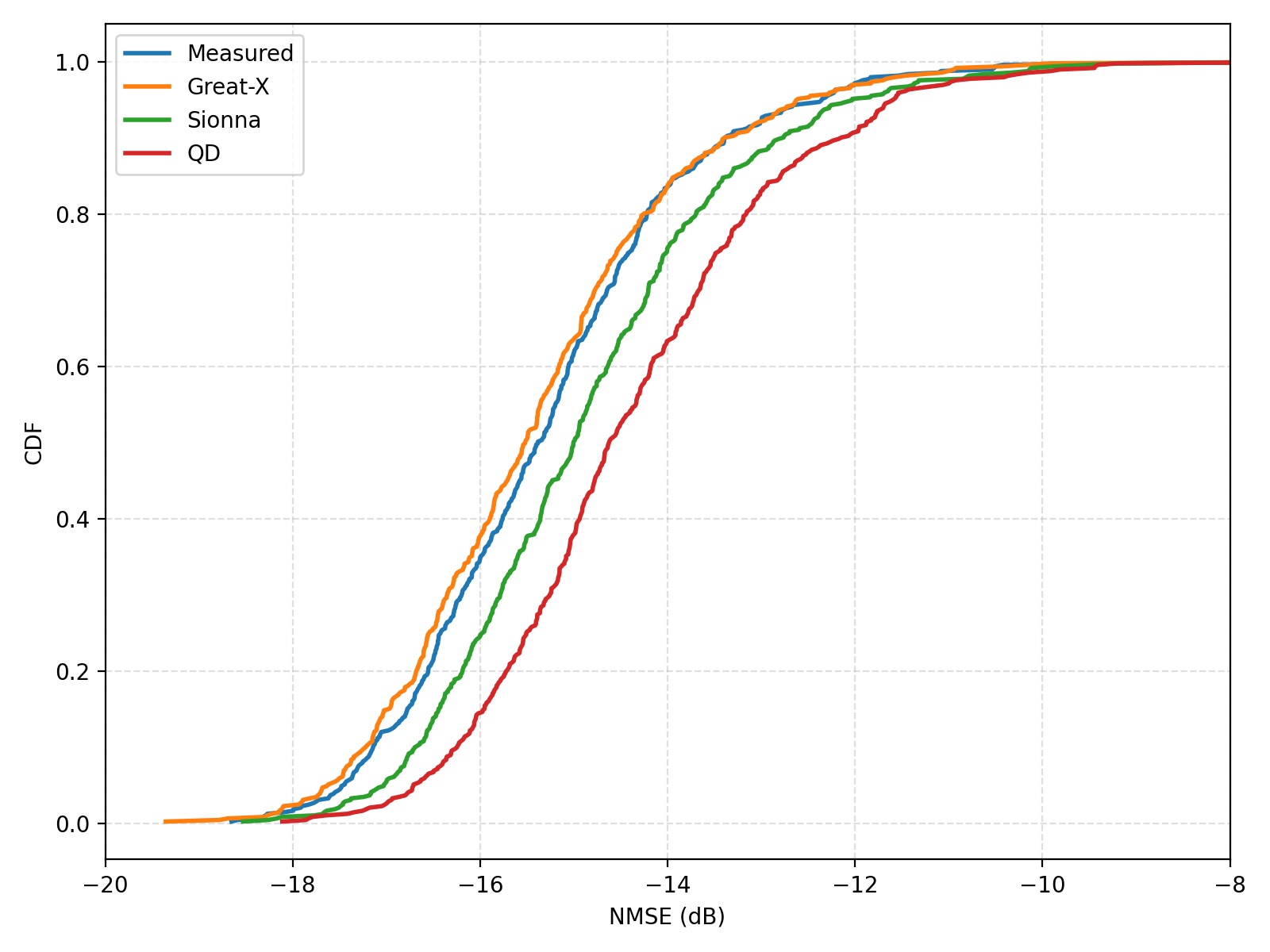}\label{fig:1_8}
}
\subfigure[Compression ratio $1/16$.]{
\includegraphics[width=0.96\columnwidth]{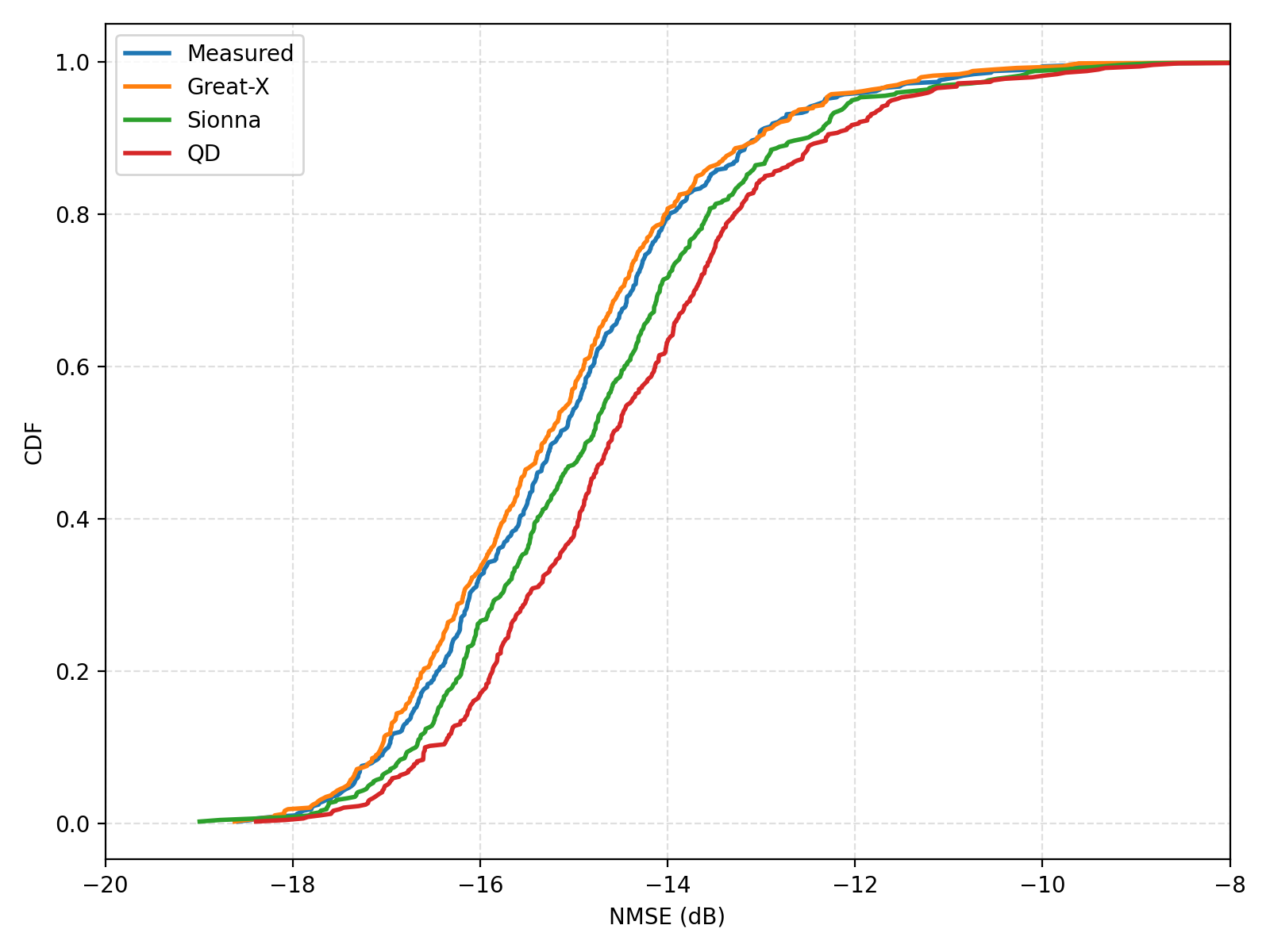}\label{fig:1_16}
}
\subfigure[Compression ratio $1/32$.]{
\includegraphics[width=0.96\columnwidth]{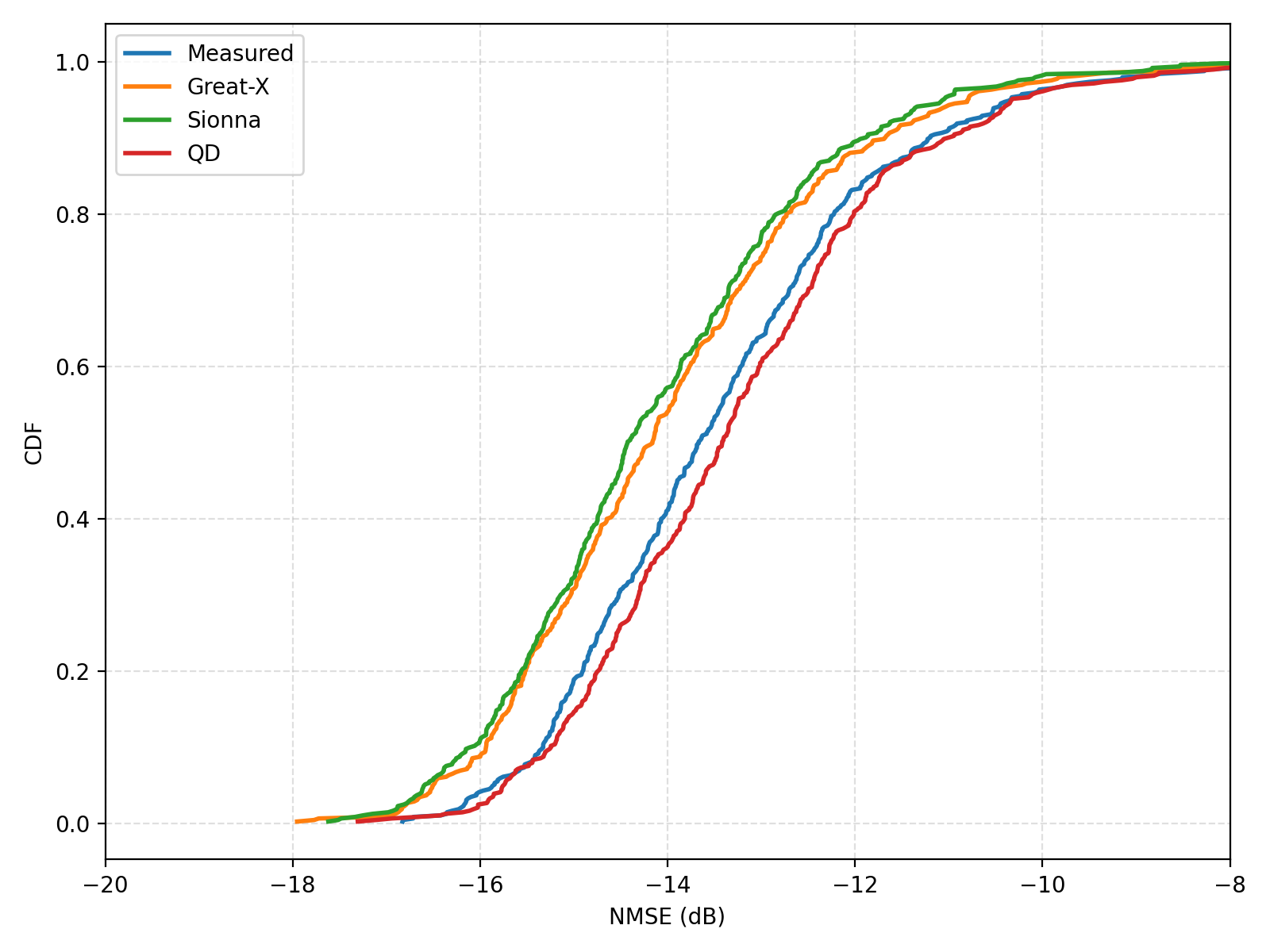}\label{fig:1_32}
}
    \caption{Cumulative Distribution Function (CDF) of CSI feedback performance (NMSE).}
    \label{fig:CSI_feedback}
\end{figure*}

\subsection{Wireless Positioning}
We develop a baseline multi-modal regression model for 3D positioning using ResNet34 \cite{he2016deep} as the backbone to support CSI, RGB, and depth inputs. Each modality is encoded independently into a 512-dimensional feature vector. Specifically, the RGB and depth images utilize a modified ResNet34 architecture where the first layer is adjusted for depth and the final fully connected (FC) layer is removed. Simultaneously, CSI tensors undergo a 2D Fast Fourier Transform (2D-FFT) and are concatenated into four channels before being processed by the adapted ResNet34. The resulting modality features $f_i \in \mathbb{R}^{512}$ are normalized via LayerNorm and fused through an attention mechanism, which employs softmax weights to compute a weighted sum across all modalities. The integrated model is trained using the Mean Squared Error (MSE) loss function to regress precise 3D coordinates.

The quantitative results presented in Table \ref{tab:kfold_summary} provide strong empirical evidence for the superiority of the proposed GREAT-X. The fundamental rationale for this strategy lies in addressing the `missing modality' challenge inherent in real-world deployment: while high-fidelity visual and geometric data are abundant within the Great-X simulation environment, such modalities are typically inaccessible in field measurements due to hardware costs and complex environmental constraints.

To bridge this gap, the proposed GREAT-X facilitates the transfer of environmental spatial awareness from a multi-modal `teacher' model—trained in the data-rich simulator—to a `student' model that relies exclusively on single-modal wireless signals during inference. As demonstrated in Table \ref{tab:kfold_summary}, the MMKD-enhanced student model achieves the best overall performance, with a mean error of $0.1598 \mathrm{m}$ and a sub-decimeter-level median error of $0.1222 \mathrm{m}$. Compared with the Great-X Baseline, whose mean error is $0.3529 \mathrm{m}$, the proposed method reduces the mean positioning error by approximately $54.7\%$. It also outperforms other simulation-based baselines, including Sionna ($0.3565 \mathrm{m}$) and QD/Quadriga ($0.4743 \mathrm{m}$), and even achieves lower error than the measured-data baseline ($0.1840 \mathrm{m}$). These results demonstrate that MMKD can effectively convert cross-modal geometric priors into robust wireless feature representations, thereby improving positioning accuracy without increasing the complexity of real-world hardware deployment.

\subsection{CSI Feedback}

To validate the physical-layer fidelity of the simulated CSI, CsiNet \cite{wen2018deep} is adopted as the representative benchmark. As a seminal deep learning-based architecture, CsiNet employs a convolutional neural network (CNN) autoencoder to achieve efficient CSI compression and reconstruction within the angular-delay domain. This foundational baseline facilitates a rigorous evaluation of whether the synthetic channel matrices retain essential structural properties, including sparsity and spatial correlation. The realism of the generated CSI features is quantitatively assessed by comparing the reconstruction performance of CsiNet, measured in terms of the Normalized Mean Squared Error (NMSE), across both simulated and real-world datasets. This comparison further validates the compatibility of the simulator with established CSI feedback algorithms.

Table \ref{tab:csi_results} validates the physical consistency and credibility of the proposed Great-X simulator by comparing its CSI reconstruction performance with established simulation benchmarks, including Sionna and QD. The credibility of the simulation data is mainly reflected in the feature transferability from Stage 1 to Stage 2. In Stage 1 (Zero-shot), the model pre-trained on Great-X data learns representative channel structures and achieves performance comparable to, or better than, models trained on other simulation datasets across different compression ratios. This indicates that Great-X can effectively capture essential multipath propagation characteristics and spatial correlations in wireless channels.

More importantly, the Stage 2 fine-tuning results show that Great-X pre-training provides a strong initialization for domain adaptation to measured data. After fine-tuning with limited measured samples, the NMSE of Great-X consistently improves to around $-15,\mathrm{dB}$ across different compression ratios. For example, at $CR=1/8$, Great-X achieves an NMSE of $-14.94 \mathrm{dB}$, outperforming Sionna ($-14.41 \mathrm{dB}$), QD ($-13.95 \mathrm{dB}$), and the Measured Only direct-training baseline ($-14.84 \mathrm{dB}$). Similar advantages can also be observed at $CR=1/4$ and $CR=1/16$, where Great-X obtains the best NMSE among all compared methods. These results suggest that the CSI samples generated by Great-X are not merely synthetic patterns, but contain physically meaningful channel characteristics that are transferable to real measured environments.

\subsection{Computational Complexity and Simulation Speed}
The major computational overhead of Great-X mainly comes from the execution of the scene construction and the synchronous advancement through \textit{world.tick()} in the Unreal Engine, rather than the CSI computation itself. Specifically, on a platform equipped with an RTX 4090D GPU and an Intel i9-14900K CPU, Great-X requires approximately 0.926 s per frame on average, where the scene synchronization and ticking process accounts for about 0.915 s, while CSI computation only takes about 0.009 s. In comparison, Sionna requires approximately 0.122 s per simulated point on average. This observation suggests that the additional runtime cost of Great-X is mainly introduced by the high-fidelity embodied simulation pipeline, including 3D scene execution, physical simulation, and frame-level synchronization, rather than by the CSI computation module itself.
\section{Conclusion}
\label{sec:conclusion}
This paper presented Great-X, a unified single-engine multi-modal simulator that overcomes the fragmentation, synchronization issues, and limited fidelity of prior simulators for 6G. By embedding a custom ray-tracing electromagnetic model directly within Unreal Engine and enforcing a deterministic fixed-step clock with physically based rendering materials co-located to electromagnetic properties, Great-X achieves pixel-level spatial consistency and zero-latency jitter across CSI, RGB, depth, LiDAR, radar, and event-camera data streams. The accompanying Great-MCD dataset supplies the research community with an unprecedented volume of high-quality, aligned multi-modal samples for low-altitude UAV scenarios. Rigorous comparisons against the DICHASUS measurement campaign confirm that Great-X reproduces realistic channel impulse responses in complex indoor environments. Downstream experiments further illustrate its practical value: a multi-modal knowledge-distillation framework trained on Great-X data surpasses baselines from Sionna, Quadriga, and even real measurements in 3D positioning accuracy, while CsiNet-based CSI feedback demonstrates strong zero-shot generalization and improved fine-tuned performance on measured data. These results collectively validate the simulator’s physical fidelity and its capacity to accelerate the development of robust, generalizable ISAC models.

\bibliographystyle{IEEEtran}
\bibliography{main}

\end{document}